\documentclass{bmvc2k}
\usepackage{booktabs}

\title{Diffusion Trajectory Modeling \\for Semantic Correspondence}

\addauthor{Yusung Choi}{cyscyb@gmail.com}{1}

\addinstitution{
 Department of Computer Engineering\\
 Pukyong National University\\
 Busan, Republic of Korea
}

\runninghead{Y. Choi}{Diffusion Trajectory Modeling}

\begin{document}

\maketitle

\begin{abstract}
Diffusion models generate images through an iterative diffusion process, and recent studies have demonstrated that the intermediate feature maps produced during this process contain rich visual representations, leading to their adoption across a variety of downstream tasks. However, most existing approaches are limited to either using a single feature map at a specific timestep or aggregating feature maps across multiple timesteps. We observe that intermediate representations in the diffusion process form meaningful trajectories along the time axis. In particular, the representation of each spatial patch evolves progressively throughout the generative process, encoding semantics that are difficult to capture from static snapshots alone. This observation motivates the need to treat diffusion representations as temporally structured trajectories rather than static snapshots. To this end, we propose Diffusion Trajectory Modeling (DTM), a framework that interprets the temporal evolution of each spatial patch as a trajectory and leverages it for semantic correspondence. By effectively modeling patch-wise trajectories generated across multiple timesteps, DTM captures correspondence cues that prior methods are not designed to capture. We further demonstrate empirically that spatially corresponding patches form similar trajectory patterns throughout the diffusion process, suggesting that the temporal axis of diffusion carries semantic information. Experiments on SPair-71k, SPair-U and AP-10K show that DTM achieves strong performance, presenting a new perspective for exploiting diffusion representations from a trajectory-centric viewpoint.

\end{abstract}

\section{Introduction}

The fundamental goal of computer vision is to extract task-relevant information from complex visual inputs and organize it into generalizable representations. From this perspective, many vision problems ultimately reduce to the challenge of learning robust and discriminative representations. Early computer vision research pursued local invariance and discriminability through handcrafted descriptors such as SIFT~\cite{lowe2004distinctive} and HOG~\cite{dalal2005histograms}, while subsequent CNN~\cite{he2016deep} and Vision Transformer-based~\cite{dosovitskiy2020image, amir2021deep} methods enabled richer and more transferable visual representations through large-scale learning. More recently, generative models~\cite{rombach2022high, ho2020denoising, tang2023emergent} — and diffusion models in particular — have emerged as a promising new source of visual representations, as they learn internal representations that capture semantic and structural information across multiple levels of abstraction. However, existing approaches to leveraging diffusion representations have largely been limited to selecting features at specific timesteps and layers~\cite{tang2023emergent, gan2025unleashing}, or aggregating features across multiple stages~\cite{zhang2023tale, luo2023diffusion}, leaving the temporally evolving nature of the diffusion process largely unexploited.

\begin{figure}[t]
\centering
\includegraphics[width=0.85\linewidth]{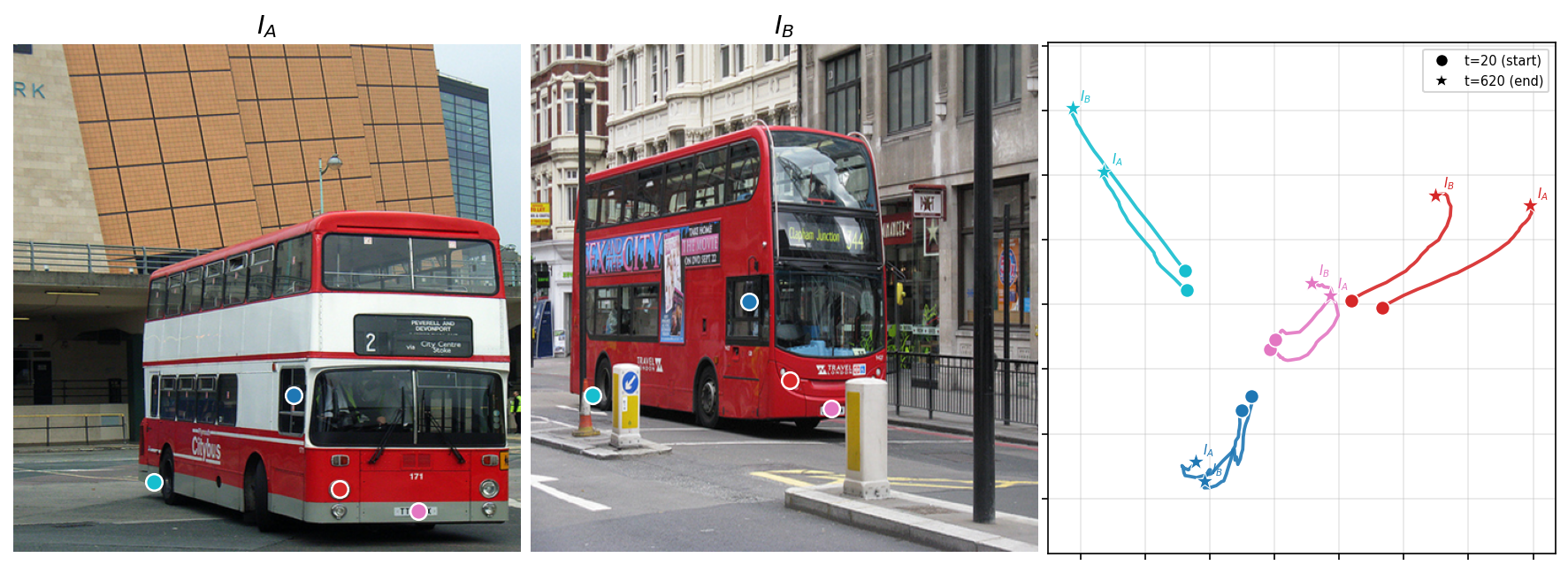}
\caption{Visualization of diffusion trajectories across image pair $I_A$ and $I_B$. Each color represents a corresponding patch pair, where circle ($\bullet$) and star ($\star$) denote features at $timestep=20$ and $timestep=620$, respectively.}
\label{fig:trajectory_teaser}
\end{figure}

In this work, we take the perspective that diffusion representations should be viewed not as static features at individual timesteps, but as trajectories along which spatial patches evolve over time. During the diffusion process, the representation of each spatial patch changes progressively along the time axis, forming not a mere collection of intermediate states, but a temporally ordered evolution of representations. Consequently, existing approaches~\cite{luo2023diffusion, gan2025unleashing, zhang2023tale, tang2023emergent} that consume diffusion representations as snapshots at a single timestep are fundamentally limited in that they discard this temporal structure.
To validate this perspective, we extract diffusion trajectories for multiple semantically corresponding patch pairs and visualize them via PCA (Figure~\ref{fig:trajectory_teaser}), where each color represents one corresponding pair. Patches that are semantically corresponding form strikingly similar trajectories throughout the diffusion process, even across different images, whereas patches of different colors exhibit clearly divergent trajectories. This suggests that the temporal evolution of representations — beyond feature similarity at any single timestep — can itself reflect semantic consistency, motivating the use of trajectories as units of representation.
To fully exploit this temporal structure, we propose \textbf{Diffusion Trajectory Modeling (DTM)}, a framework that directly models patch-wise trajectories for semantic correspondence. DTM encodes the trajectory of each spatial patch as a sequential signal, incorporating temporal variation patterns into the representation that would be difficult to capture from static snapshots alone. Through this design, DTM effectively captures correspondence cues that prior methods~\cite{luo2023diffusion, zhang2023tale, gan2025unleashing, tang2023emergent} are not explicitly designed to capture.
We evaluate DTM on three semantic correspondence benchmarks: SPair-71k~\cite{min2019spair}, SPair-U~\cite{mariotti2025jamais} and AP-10K~\cite{yu2021ap}. Results show that trajectory information improves correspondence performance over existing representations.
\begin{itemize}
\item We introduce a trajectory-centric perspective that interprets diffusion representations as temporally structured trajectories and propose DTM, a framework that leverages this view for semantic correspondence.
\item We provide empirical evidence that semantically corresponding spatial patches form similar trajectories throughout the diffusion process, suggesting that trajectories serve as structural units encoding semantic information.
\item We demonstrate through experiments on SPair-71k~\cite{min2019spair}, SPair-U~\cite{mariotti2025jamais} and AP-10K~\cite{yu2021ap} that DTM improves correspondence performance over existing baselines.
\end{itemize}

\section{Related Work}
\textbf{Diffusion Trajectory.}
Diffusion models can be understood as trajectory-based generative processes~\cite{ho2020denoising,song2020score} that progressively transition from a noisy state to a data state. Existing works have formalized the diffusion process from the perspectives of stochastic differential equations (SDEs~\cite{song2020score}), probability flow ODEs~\cite{song2020score}, and reverse dynamics~\cite{song2020denoising}, interpreting the generative process as a continuous path in state space. More recent studies have analyzed the geometric structure and sampling dynamics of diffusion trajectories, or leveraged them to improve sampling efficiency~\cite{lu2022dpm,zhang2022fast}. However, these works primarily focus on understanding and controlling the generative process, with analyses largely focused on sample-level or latent-level dynamics. In contrast, how spatial patches evolve across timesteps throughout the diffusion process — and how such patch-level trajectories can be exploited as structured representations for downstream vision tasks — remains largely unexplored. Our work is distinguished from prior work in that we approach diffusion trajectories from the perspective of patch-level representation dynamics rather than generative dynamics.

\vspace{0.3em}
\noindent\textbf{Representation for Semantic Correspondence.}
The quality of representations is a critical factor determining correspondence performance in semantic matching. Self-supervised representations such as DINO~\cite{caron2021emerging, oquab2023dinov2} have served as strong baselines, and it has since been shown~\cite{tang2023emergent} that the internal representations of diffusion models can also serve as effective descriptors for semantic correspondence. DIFT~\cite{tang2023emergent} demonstrated that diffusion features alone can achieve strong correspondence performance, while SD+DINO~\cite{zhang2023tale} further improved zero-shot semantic correspondence by combining the complementary strengths of Stable Diffusion~\cite{rombach2022high} and DINO~\cite{oquab2023dinov2}. Diffusion Hyperfeatures~\cite{luo2023diffusion} consolidated multi-scale and multi-timestep feature maps into per-pixel descriptors via a learnable aggregation network. More recently, DiTF~\cite{gan2025unleashing} showed that semantically discriminative features can be effectively extracted from Diffusion Transformers as well. Nevertheless, these works~\cite{tang2023emergent,gan2025unleashing,zhang2023tale,luo2023diffusion} primarily focus on selecting features at specific timesteps or combining multiple representations, and attempts to directly incorporate the temporal structure inherent to the diffusion process into correspondence representations remain limited.

\section{Diffusion Trajectory}
We first define the notion of diffusion trajectory as used in this work, and provide an analysis suggesting that semantically corresponding spatial patches exhibit similar trajectory patterns throughout the diffusion process. Building on this observation, we describe DTM, our framework for leveraging patch-wise trajectories for semantic correspondence.

\subsection{Diffusion Trajectory Definition}
\label{sec:3_1}
Diffusion models encode an input image into a latent space and process it iteratively across multiple timesteps, producing rich internal feature maps at each block.
We adopt a pre-trained Diffusion Transformer as our backbone and extract features following the DiTF framework~\cite{gan2025unleashing}.
Specifically, the noisy latent obtained at timestep $t$ is passed through a sequence of DiT blocks, yielding an intermediate feature $z_t^k$ at the $k$-th block.
To address the massive activations problem in DiT features --- a small number of feature dimensions with disproportionately large activation values that degrade correspondence performance based on cosine similarity --- DiTF applies AdaLN-based channel-wise modulation to this pre-AdaLN feature:
\begin{equation}
    \hat{z}_t^k = (1+\gamma_k)\mathrm{LayerNorm}(z_t^k) + \beta_k
\end{equation}
where $\gamma_k, \beta_k$ are channel-wise scale and shift parameters regressed by an MLP conditioned on the timestep $t$ and text embedding $c$.
A channel discard strategy is further applied to eliminate residual weak activations.
Through this process, we obtain a semantically discriminative feature representation $\hat{z}_t^k$ for each spatial patch at a given timestep $t$ and block $k$.

The feature map $\hat{z}_t^k \in \mathbf{R}^{C \times H \times W}$ obtained above represents the full spatial feature map of the input image at timestep $t$ and block $k$, serving as the fundamental building block for constructing diffusion trajectories. Hereafter, we fix the block index $k$ and write $\hat{z}_t$ for brevity.

Given an input image, we extract feature maps at $N$ selected timesteps $\mathcal{T} = \{t_1, t_2, \dots, t_N\}$, yielding $\hat{z}_{t_i} \in \mathbf{R}^{C \times H \times W}$ at each timestep $t_i$.
To decompose this feature map into patch-level representations, we index spatial positions over the $H \times W$ grid, where each position $p \in \{1, \dots, H \times W\}$ corresponds to a distinct spatial patch.
The feature vector of the patch at position $p$ is extracted as:
\begin{equation}
    \hat{z}_{t_i}^p = \hat{z}_{t_i}[p] \in \mathbf{R}^{C}
\end{equation}
where $\hat{z}_{t_i}[p]$ denotes the channel vector at spatial position $p$ of the feature map $\hat{z}_{t_i}$.
Arranging the patch features across all timesteps $t_i \in \mathcal{T}$ in temporal order, we define the \textbf{diffusion trajectory} of patch $p$ as:
\begin{equation}
    \tau^p = \left[ \hat{z}_{t_1}^p,\, \hat{z}_{t_2}^p,\, \dots,\, \hat{z}_{t_N}^p \right] \in \mathbf{R}^{N \times C}
\end{equation}
Rather than treating each timestep feature independently, we view $\tau^p$ as a temporally ordered sequence, regarding the evolution of patch representations across timesteps as a structured signal.
In its raw form, $\tau^p$ is a stacked sequence of per-timestep features arranged in temporal order; nevertheless, we argue that this temporal ordering itself carries meaningful information.
We first analyze in Section~\ref{sec:3_2} whether this temporal ordering reflects semantic information.
Building on this analysis, we show that explicitly modeling $\tau^p$ enables the capture of semantic dynamics that are difficult to recover from single-timestep features alone, and describe the specific modeling approach in Section~\ref{sec:3_3}.

\subsection{Trajectory Similarity Analysis}
\label{sec:3_2}

\begin{figure}[htbp]
\centering
\includegraphics[width=0.85\linewidth]{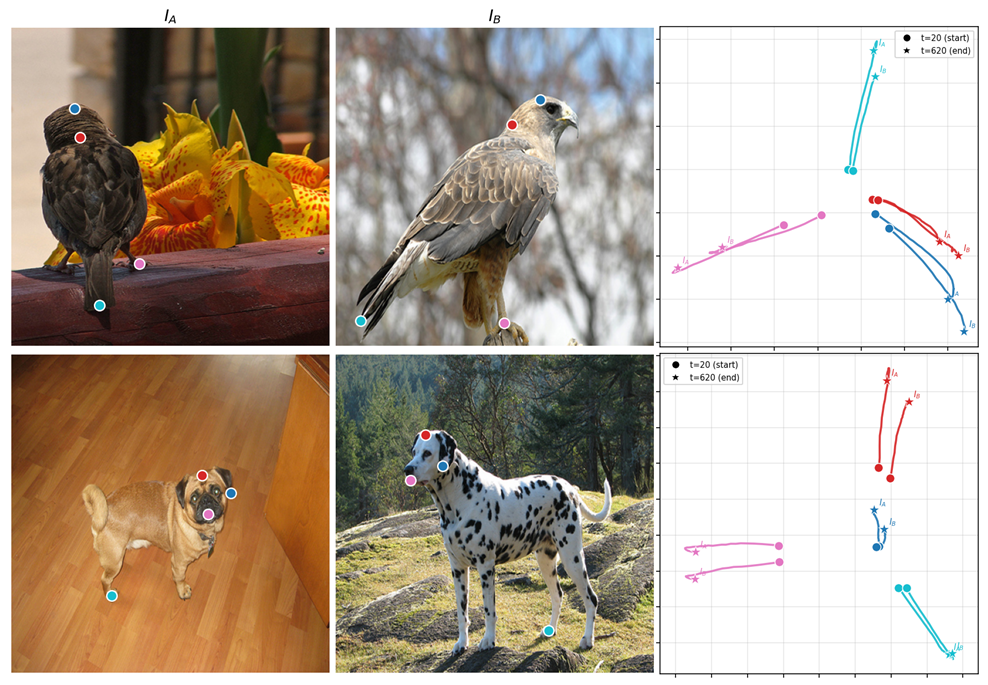}
\caption{Diffusion trajectory visualization of semantically corresponding patch pairs.}
\label{fig:trajectory_main}
\end{figure}

To analyze whether the trajectory $\tau^p$ defined in Section~\ref{sec:3_1} genuinely reflects semantic correspondence, we conduct a trajectory analysis on patch pairs extracted from images $I_a$ and $I_b$ (Figure~\ref{fig:trajectory_main}).

Specifically, leveraging keypoint annotations from SPair-71k~\cite{min2019spair}, we select $M$ semantically corresponding patch pairs $\{(p_a^m, p_b^m)\}_{m=1}^{M}$ across two images and extract their respective trajectories $\tau^{p_a^m}, \tau^{p_b^m} \in \mathbf{R}^{N \times C}$.
The total $2M$ trajectories extracted from both images are then jointly projected onto a shared low-dimensional space via PCA.
Applying PCA jointly ensures that trajectories from both images are compared within the same basis space, enabling a meaningful geometric comparison.
In the visualization, the two trajectories belonging to the same corresponding pair $(p_a^m, p_b^m)$ are rendered in the same color, while different corresponding pairs are assigned distinct colors.

Figure~\ref{fig:trajectory_main} presents the visualization results across multiple image pairs.
Notably, trajectories of corresponding patch pairs, shown in matching colors, form remarkably similar patterns in the shared PCA space, despite being extracted from different images.
In contrast, non-corresponding patch pairs exhibit clearly distinct and separable patterns.
Furthermore, the PCA visualization suggests that temporal variation across the entire trajectory provides correspondence cues beyond individual timestep features.
Semantically corresponding patches are similar not only at individual timesteps, but also in how their representations evolve throughout the diffusion process, indicating that temporal evolution itself may reflect semantic structure.

These observations support two important conclusions.
First, diffusion trajectories can function as signals that reflect semantic correspondence across images.
Second, this temporal structure is difficult to capture through existing approaches~\cite{tang2023emergent,gan2025unleashing,zhang2023tale,luo2023diffusion} that select features at specific timesteps or aggregate them across timesteps, highlighting the need for an approach that models the entire trajectory as a sequential signal.
Building on these findings, Section~\ref{sec:3_3} describes our specific method for explicitly modeling $\tau^p$ and leveraging it for semantic correspondence.

\subsection{Diffusion Trajectory Modeling}
\label{sec:3_3}
Given the trajectory $\tau^p$ extracted from the diffusion process, the most straightforward approach would be to directly use the raw concatenation of per-timestep features.
However, this is problematic for two reasons: the dimensionality of the representation becomes excessively large as the trajectory spans multiple timesteps, and the naive concatenation structure fails to capture the sequential and structural characteristics inherent to the trajectory.
To address this, we propose \textbf{Diffusion Trajectory Modeling (DTM)}, a framework that treats the trajectory as a sequence and transforms it into a representation suitable for semantic correspondence.

DTM treats each patch trajectory as a sequential signal and builds upon State Space Models (SSMs)~\cite{gu2021efficiently, gu2023mamba}, which are well-suited for modeling long-range dependencies and sequential structure.
In particular, we leverage Mamba~\cite{gu2023mamba}, which exploits a selective state space mechanism to efficiently incorporate information across the entire sequence.

\begin{figure}[t]
\centering
\includegraphics[width=\linewidth]{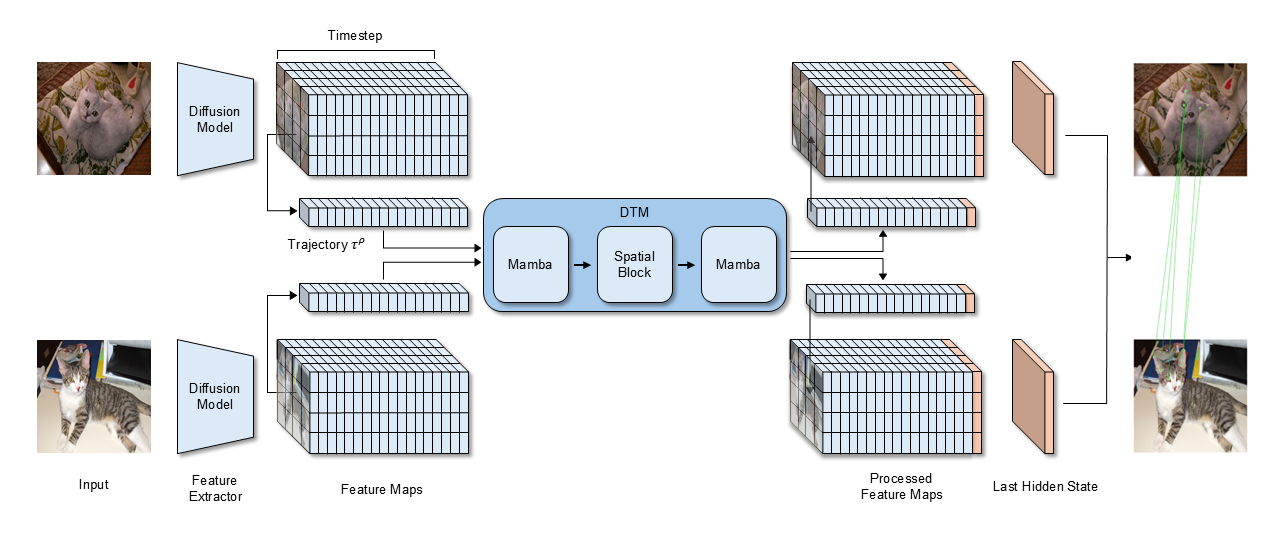}
\caption{Overview of the proposed Diffusion Trajectory Modeling (DTM) framework.}
\label{fig:DTM}
\end{figure}

Specifically, the trajectory $\tau^p \in \mathbf{R}^{N \times C}$ of each spatial patch $p$ is processed independently by a unidirectional Mamba encoder $\mathcal{M}_1$, which operates along the timestep dimension with sequence length $N$ and $C$-dimensional patch features as tokens.
The outputs of $\mathcal{M}_1$ across all spatial patches are then collectively reshaped into a spatiotemporal feature volume of size $N \times H \times W \times C$.
A shared spatial block consisting of residual convolutional layers is then applied to each timestep-wise feature map independently, incorporating local spatial context among neighboring patches while preserving the temporal dimension.
The spatially enriched features are subsequently redistributed back into per-patch sequences and passed through a second unidirectional Mamba encoder $\mathcal{M}_2$, which further models the temporal structure with the benefit of spatial context.
As a unidirectional sequential model, Mamba accumulates information across the entire input sequence into its final hidden state; we therefore take the last hidden state of $\mathcal{M}_2$ as the trajectory-aware representation of the patch:
\begin{equation}
    h^p = \mathcal{M}_2(\text{SpatialBlock}(\mathcal{M}_1(\tau^p))) \in \mathbf{R}^{d}
\end{equation}
This interleaving of temporal encoding and spatial context integration goes beyond simply applying a sequence model, jointly capturing the temporal evolution of each patch and its complementary spatial relationships with neighboring patches.
Here, the last hidden state $h^p$ is a representation that implicitly encodes the temporal evolution of patch $p$ throughout the diffusion process, capturing correspondence cues that are difficult to recover from single-timestep features alone.
Given the representations $h^{p_a}$ and $h^{p_b}$ extracted for patches in images $I_a$ and $I_b$, correspondence is estimated based on cosine similarity:
\begin{equation}
    \text{sim}(p_a, p_b) = \frac{h^{p_a} \cdot h^{p_b}}{\|h^{p_a}\| \|h^{p_b}\|}
\end{equation}
The Mamba encoder is trained using keypoint annotations as supervision.
A bidirectional InfoNCE loss~\cite{oord2018representation} is employed to encourage representations of corresponding patch pairs to be similar while pushing non-corresponding pairs apart:
\begin{equation}
    \mathcal{L} = \mathcal{L}_{\text{InfoNCE}}(h^{p_a} \rightarrow h^{p_b}) + \mathcal{L}_{\text{InfoNCE}}(h^{p_b} \rightarrow h^{p_a})
\end{equation}
The effectiveness of the proposed DTM is validated through the experiments.

\section{Experiments}
\subsection{Experimental Details}

Trajectories are extracted from the pre-trained FLUX.1-dev~\cite{flux2024} and fed into DTM.
For SPair-71k~\cite{min2019spair} and SPair-U~\cite{mariotti2025jamais}, we evaluate both benchmarks using the same checkpoint trained on the training split of SPair-71k.
For AP-10K~\cite{yu2021ap}, we train a separate checkpoint on its training split and evaluate under the Intra-Species, Cross-Species, and Cross-Family settings.
For all experiments, we use the AdamW optimizer~\cite{loshchilov2017decoupled} with a cosine learning rate scheduler, decaying the learning rate from $1\times10^{-4}$ to $1\times10^{-5}$ over a total of 4,000 training steps.
Input images are processed at a resolution of $960\times960$, and all experiments are conducted on a single NVIDIA A100 SXM GPU.

For trajectory construction, we sample timesteps from the full $T=1000$ diffusion schedule at intervals of 20 within the range $t \in \{20, 40, 60, \dots, 620\}$, yielding a total of $N=31$ timesteps per trajectory, and fix the block index at $k=28$ following DiTF~\cite{gan2025unleashing}.
Timesteps beyond $T=620$ are excluded as the corresponding feature maps contain excessive noise that disrupts the semantic structure of the trajectory.

\subsection{Datasets}

\noindent\textbf{SPair-71k}~\cite{min2019spair} is a large-scale semantic correspondence benchmark comprising 70,958 image pairs across 18 object categories. It evaluates correspondence between diverse object instances within the same category, providing a challenging setting with substantial variations in viewpoint, scale, and appearance. Its large scale and categorical diversity enable a thorough assessment of the generalization ability of correspondence methods.

\vspace{0.3em}
\noindent\textbf{SPair-U}~\cite{mariotti2025jamais} is an extension of SPair-71k that introduces novel keypoint annotations not seen during training, designed to expose the generalization gap in supervised semantic correspondence methods. While SPair-71k evaluates performance on standard annotated keypoints, SPair-U assesses whether methods can generalize beyond sparsely annotated training keypoints, providing a more rigorous evaluation of correspondence generalization ability.

\vspace{0.3em}
\noindent\textbf{AP-10K}~\cite{yu2021ap} is a benchmark comprising animal images captured in the wild, originally proposed for animal pose estimation.
For semantic correspondence evaluation, the benchmark is decomposed into Intra-Species (IS), which evaluates correspondence within the same species as those used for training; Cross-Species (CS), which evaluates correspondence to unseen species within the same family; and Cross-Family (CF), which evaluates correspondence to unseen families.
This decomposition enables the generalization ability of correspondence methods to be assessed as the domain gap progressively widens.

\subsection{Metrics}
\label{sec:metrics}

We evaluate using Percentage of Correct Keypoints (PCK), the standard metric for semantic correspondence. A predicted keypoint is considered correct if it falls within a radius of $\alpha \cdot \max(h, w)$ from the ground-truth keypoint, where $h$ and $w$ refer to the height and width of the object bounding box ($\alpha_\text{bbox}$).

\begin{table*}[htbp]
\centering
\setlength{\tabcolsep}{3pt}
\renewcommand{\arraystretch}{1.35}
\scriptsize
\begin{tabular*}{\textwidth}{@{\extracolsep{\fill}}l|ccc|ccc|ccc|ccc|ccc@{}}
\toprule
\multicolumn{1}{l}{} & \multicolumn{3}{c}{SPair-71k} & \multicolumn{3}{c}{SPair-U} & \multicolumn{3}{c}{AP-10K (IS)} & \multicolumn{3}{c}{AP-10K (CS)} & \multicolumn{3}{c}{AP-10K (CF)} \\
\cmidrule(lr){2-4}\cmidrule(lr){5-7}\cmidrule(lr){8-10}\cmidrule(lr){11-13}\cmidrule(lr){14-16}
\multicolumn{1}{l}{Method} & \multicolumn{3}{c}{$\alpha$: bbox} & \multicolumn{3}{c}{$\alpha$: bbox} & \multicolumn{3}{c}{$\alpha$: bbox} & \multicolumn{3}{c}{$\alpha$: bbox} & \multicolumn{3}{c}{$\alpha$: bbox} \\
\multicolumn{1}{l}{} & \multicolumn{3}{c}{0.05 \quad 0.1 \quad 0.15} & \multicolumn{3}{c}{0.05 \quad 0.1 \quad 0.15} & \multicolumn{3}{c}{0.01 \quad 0.05 \quad 0.10} & \multicolumn{3}{c}{0.01 \quad 0.05 \quad 0.10} & \multicolumn{3}{c}{0.01 \quad 0.05 \quad 0.10} \\
\midrule
Zero-shot                        & 49.5 & 61.1 & 67.2 & 37.1 & 53.4 & 63.0 & 7.6 & 49.7 & 62.2 & 6.8 & 48.2 & 61.7 & 5.7 & 34.2 & 48.9 \\
GAP                              & 52.0 & 64.1 & 70.3 & \underline{38.0} & \underline{54.9} & \underline{65.3} & 8.2 & 51.5 & 64.2 & 7.7 & 50.6 & 63.0 & 6.5 & 36.4 & 50.7 \\
Per-pixel descriptors            & \underline{65.2} & \underline{75.1} & \underline{79.3} & 36.0 & 54.6 & 64.6 & \underline{17.0} & \underline{65.3} & \underline{81.4} & \underline{15.2} & \underline{64.1} & \underline{79.7} & \underline{12.8} & \underline{56.8} & \underline{71.5} \\
\textbf{DTM (Ours)}              & \textbf{70.0} & \textbf{80.5} & \textbf{84.4} & \textbf{39.2} & \textbf{58.7} & \textbf{68.3} & \textbf{20.5} & \textbf{71.0} & \textbf{86.2} & \textbf{18.9} & \textbf{68.2} & \textbf{84.8} & \textbf{15.3} & \textbf{62.1} & \textbf{77.9} \\
\bottomrule
\end{tabular*}
\vspace{0.3em}
\caption{Per-image PCK comparison on SPair-71k, SPair-U, and AP-10K. All methods share the same FLUX backbone and differ only in how diffusion features are utilized. We compare a single-timestep feature map baseline (Zero-shot), a multi-timestep Global Average Pooling baseline (GAP), a reimplementation of Diffusion Hyperfeatures~\cite{luo2023diffusion} that integrates multi-timestep feature maps via a learnable weighted aggregation (per-pixel descriptors), and the proposed DTM. Per-pixel feature descriptors and DTM are trained under identical settings and differ only in their architecture. AP-10K results are reported under Intra-Species (IS), Cross-Species (CS), and Cross-Family (CF) settings, spanning increasing domain gaps. The highest PCK is highlighted in \textbf{bold}, and the second highest is \underline{underlined}.}
\label{tab:main_comparison}
\end{table*}

\begin{table*}[htbp]
\centering
\setlength{\tabcolsep}{3pt}
\scriptsize
\resizebox{\textwidth}{!}{%
\begin{tabular}{l|cccccccccccccccccc|c}
\hline
Method & aero & bike & bird & boat & bottle & bus & car & cat & chair & cow & dog & horse & motor & person & plant & sheep & train & tv & All \\
\hline

DIFT~\cite{tang2023emergent}          & 63.5 & 54.5 & 80.8 & 34.5 & 46.2 & 52.7 & 48.3 & 77.7 & 39.0 & 76.0 & 54.9 & 61.3 & 53.3 & 46.0 & 57.8 & 57.1 & 71.1 & 63.4 & 57.7 \\
DINOv2~\cite{oquab2023dinov2}        & 72.7 & 62.0 & 85.2 & 41.3 & 40.4 & 52.3 & 51.5 & 71.1 & 36.2 & 67.1 & 64.6 & 67.6 & 61.0 & 68.2 & 30.7 & 62.0 & 54.3 & 24.2 & 55.6\\
SD+DINO~\cite{zhang2023tale}       & 73.0 & 64.1 & 86.4 & 40.7 & 52.9 & 55.0 & 53.8 & 78.6 & 45.5 & 77.3 & 64.7 & 69.7 & 63.3 & 69.2 & 58.4 & 67.6 & 66.2 & 53.5 & 64.0 \\
DiTF (FLUX)~\cite{gan2025unleashing}          & 74.3 & 65.0 & 88.1 & 48.1 & 53.2 & 60.7 & 60.7 & 84.9 & 42.4 & 82.8 & 68.4 & 72.1 & 70.9 & 74.2 & 62.1 & 72.6 & 66.0 & 60.3 & 67.1 \\

\hline

CATs~\cite{cho2021cats}            & 52.0 & 34.7 & 72.2 & 34.3 & 49.9 & 57.5 & 43.6 & 66.5 & 24.4 & 63.2 & 56.5 & 52.0 & 42.6 & 41.7 & 43.0 & 33.6 & 72.6 & 58.0 & 49.9 \\
CATs++~\cite{cho2022cats++}          & 60.6 & 46.9 & 82.5 & 41.6 & 56.8 & 64.9 & 50.4 & 72.8 & 29.2 & 75.8 & 65.4 & 62.5 & 50.9 & 56.1 & 54.8 & 48.2 & 80.9 & 74.9 & 59.9 \\
DHF~\cite{luo2023diffusion}          & 74.0 & 61.0 & 87.2 & 40.7 & 47.8 & 70.0 & 74.4 & 80.9 & 38.5 & 76.1 & 60.9 & 66.8 & 66.6 & 70.3 & 58.0 & 54.3 & 87.4 & 60.3 & 64.9 \\
SD+DINO (S)~\cite{zhang2023tale}  & \underline{81.2} & 66.9 & \underline{91.6} & 61.4 & 57.4 & 85.3 & \textbf{83.1} & \underline{90.8} & 54.5 & \underline{88.5} & \underline{75.1} & \underline{80.2} & \underline{71.9} & 77.9 & 60.7 & \underline{68.9} & \underline{92.4} & 65.8 & 74.6\\
SD4Match~\cite{li2024sd4match}     & 75.3 & \underline{67.4} & 85.7 & \underline{64.7} & \underline{62.9} & \underline{86.6} & 76.5 & 82.6 & \underline{64.8} & 86.7 & 73.0 & 78.9 & 70.9 & \underline{78.3} & \underline{66.8} & 64.8 & 91.5 & \underline{86.6} & \underline{75.5}\\
\textbf{DTM (Ours)} & \textbf{85.0} & \textbf{72.3} & \textbf{92.2} & \textbf{71.3} & \textbf{63.2} & \textbf{88.6} & \underline{78.9} & \textbf{94.9} & \textbf{66.7} & \textbf{90.2} & \textbf{77.1} & \textbf{81.6} & \textbf{80.7} & \textbf{84.9} & \textbf{70.8} & \textbf{73.0} & \textbf{94.9} & \textbf{89.7} & \textbf{80.9} \\
\hline
\end{tabular}
}
\vspace{0.3em}
\caption{Evaluation on SPair-71k. Per-category PCK results at $\alpha_\text{bbox}=0.1$ on SPair-71k. Due to inconsistencies in evaluation protocols across prior works, we report per-keypoint PCK for zero-shot methods and per-image PCK for supervised methods separately~\cite{zhang2024telling}.}
\label{tab:spair71k}
\normalsize
\end{table*}

\smallskip
Table~\ref{tab:main_comparison} controls for a fixed backbone and training setup while varying only how diffusion features are utilized, whereas Table~\ref{tab:spair71k} provides a comparison with prior methods.

As shown in Table~\ref{tab:main_comparison}, the proposed DTM achieves the highest performance on SPair-71k and SPair-U, recording 80.5 PCK@0.1 and 58.7 PCK@0.1, respectively, outperforming the single-timestep baseline (Zero-shot), the simple aggregation baseline (GAP), and the per-pixel descriptor approach.
Zero-shot does not leverage any temporal structure of the diffusion process, as it relies solely on a single-timestep feature. GAP aggregates features across multiple timesteps, which provides richer information than a single timestep; however, averaging across timesteps suppresses timestep-specific information, limiting its ability to capture the sequential structure of the trajectory.
Per-pixel descriptors have the advantage of adaptively reflecting the importance of each feature map via a learnable weighted aggregation; however, it treats all feature maps independently and thus disregards the sequential dependencies between timesteps.
In particular, DTM consistently outperforms per-pixel descriptor across all benchmarks under identical training settings, suggesting that explicitly modeling the sequential structure of trajectories enables richer representation learning than simple weighted aggregation.

This tendency becomes even more pronounced on SPair-U, which evaluates generalization to unseen keypoints not observed during training, where the performance ranking follows DTM > GAP > Per-pixel > Zero-shot. This suggests that Per-pixel descriptor, which learns a weighted aggregation optimized for training keypoints, may overfit to the seen keypoints and struggles to generalize to unseen ones. In contrast, DTM, which models the temporal structure of trajectories, appears to learn representations that are less dependent on specific keypoints, exhibiting stronger generalization ability. A similar trend is observed on AP-10K, where DTM maintains consistently superior performance over the baselines across the IS, CS, and CF settings as the domain gap widens, demonstrating that trajectory-based representations are robust to cross-domain generalization.

Table~\ref{tab:spair71k} presents per-category PCK results on SPair-71k. Among supervised methods, DTM achieves the highest overall mean PCK of 80.9, surpassing CATs~\cite{cho2021cats} (49.9), CATs++~\cite{cho2022cats++} (59.9), DHF~\cite{luo2023diffusion} (64.9), SD+DINO~\cite{zhang2023tale} (S) (74.6), and SD4Match~\cite{li2024sd4match} (75.5).
Examining per-category results, DTM achieves notably high performance on categories with consistent global structure and distinctive semantic layout, such as cat (94.9), bird (92.2), and train (94.9). These categories tend to form stable and consistent trajectory patterns throughout the diffusion process, allowing DTM to effectively capture correspondence cues.
In contrast, relatively lower performance is observed on categories such as bottle (63.2) and chair (66.7); we speculate that large intra-class appearance and structural variations make it difficult to form consistent trajectory patterns across instances.

\subsection{Ablation}

\noindent\textbf{Backbone Ablation.} To investigate whether the effectiveness of leveraging diffusion trajectories is specific to a particular backbone or reflects a general property of diffusion models, we conduct experiments using Stable Diffusion 3.0~\cite{esser2024scaling} (SD3.0), Stable Diffusion 3.5~\cite{esser2024scaling} (SD3.5), and Stable Diffusion 1.5~\cite{rombach2022high} (SD1.5) as alternative backbones in addition to FLUX.1-dev.
Among these, SD1.5 is based on a UNet architecture, while the others belong to the DiT family.
All experiments are conducted under identical training settings and evaluated on SPair-71k.
As shown in Table~\ref{tab:backbone_ablation}, DTM remains effective across all backbones despite these architectural differences.
This suggests that the phenomenon whereby semantically corresponding patches form similar trajectories is not specific to a particular diffusion model, but rather reflects a structural property that generalizes across diffusion models.
In other words, these results support the view that leveraging diffusion trajectories as representations constitutes a backbone-agnostic approach that does not rely on the characteristics of any specific model.

\begin{table}[h]
\centering
\setlength{\tabcolsep}{18pt}
\begin{tabular}{l|c}
\hline
Backbone & PCK ($\alpha_{\mathrm{bbox}}=0.1$) \\
\hline
SD1.5       & 62.3 \\
SD3.0       & 73.1 \\
SD3.5       & 77.6 \\
FLUX.1-dev  & 80.5 \\
\hline
\end{tabular}
\vspace{0.3em}
\caption{Backbone ablation on SPair-71k ($\alpha_\text{bbox}=0.1$).}
\label{tab:backbone_ablation}
\end{table}

\noindent\textbf{Temporal order shuffle.} In Section~\ref{sec:3_2}, we observed that semantically corresponding patches form similar trajectory patterns throughout the diffusion process. To examine whether these patterns depend on the semantic chronology of the diffusion process --- that is, the temporally structured progression of noise --- we conduct two temporal order perturbation experiments.

The first applies the same random permutation to all images (Same Shuffle), preserving the alignment of temporal axes across images while disrupting the original semantic chronology of the diffusion process. The second applies a different random permutation independently to each image (Different Shuffle), such that even the alignment of temporal axes across images is no longer guaranteed. For both shuffle variants, the permutation is applied consistently during both training and evaluation.

As shown in Table~\ref{tab:temporal_shuffle}, the improvement from Different Shuffle to Same Shuffle (+2.1) reflects the contribution of temporal alignment, while that from Same Shuffle to the original order (+4.3) reflects the contribution of semantic chronology. Prior theoretical work~\cite{murphy2018janossy} suggests that training a permutation-sensitive model with random permutations approximates a permutation-invariant function; indeed, Different Shuffle (74.1), which contains no order information at all, has the same capacity as the original DTM but converges to a level similar to that of per-pixel descriptor (75.1, Table~\ref{tab:main_comparison}), a simpler learned aggregation architecture. This suggests that a performance ceiling exists for order-agnostic approaches regardless of capacity, and that the improvement of the original-order DTM beyond this ceiling stems from exploiting order information in training, rather than from capacity.

Taken together, these results empirically demonstrate that multi-timestep coverage, temporal alignment, and semantic chronology each contribute independently to the performance of diffusion trajectories, with the contribution of chronology (+4.3) being more than twice as large as that of alignment (+2.1). This supports the view that the temporal ordering of diffusion trajectories serves as a meaningful signal reflecting semantic correspondence, beyond simply being a sequential structure.

\begin{table}[h]
\centering
\setlength{\tabcolsep}{18pt}
\begin{tabular}{l|c}
\hline
Method & PCK ($\alpha_{\mathrm{bbox}}=0.1$) \\
\hline
DTM (Different Shuffle) & 74.1 \\
DTM (Same Shuffle)      & 76.2 \\
DTM (Original Order)    & 80.5 \\
\hline
\end{tabular}
\vspace{0.3em}
\caption{Temporal order shuffle ablation on SPair-71k ($\alpha_\text{bbox}=0.1$).}
\label{tab:temporal_shuffle}
\end{table}

\subsection{Qualitative Comparison}

\begin{figure}[htbp]
\centering
\includegraphics[width=0.65\linewidth]{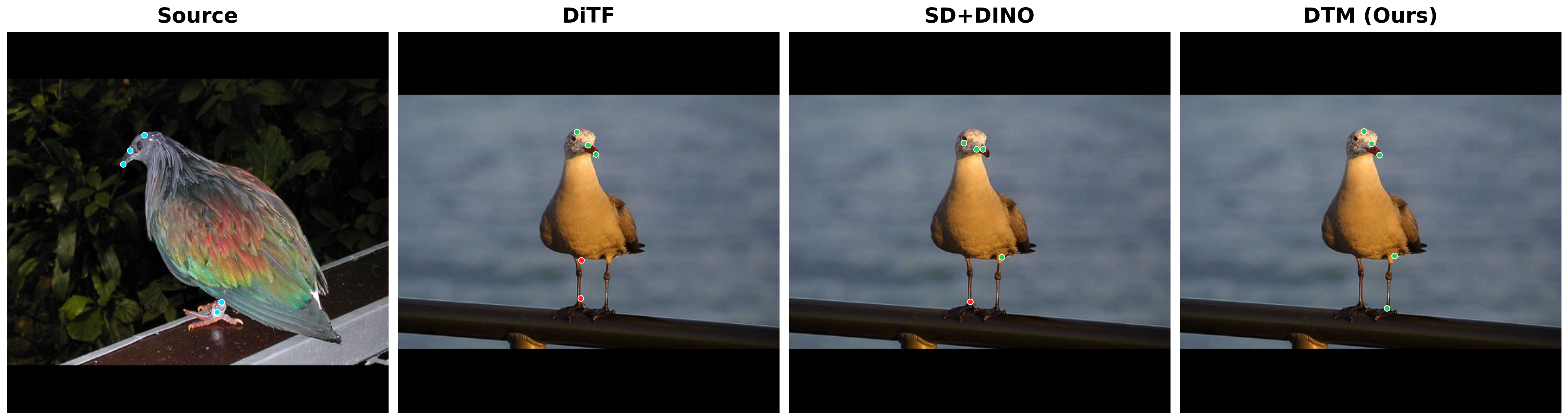}
\includegraphics[width=0.65\linewidth]{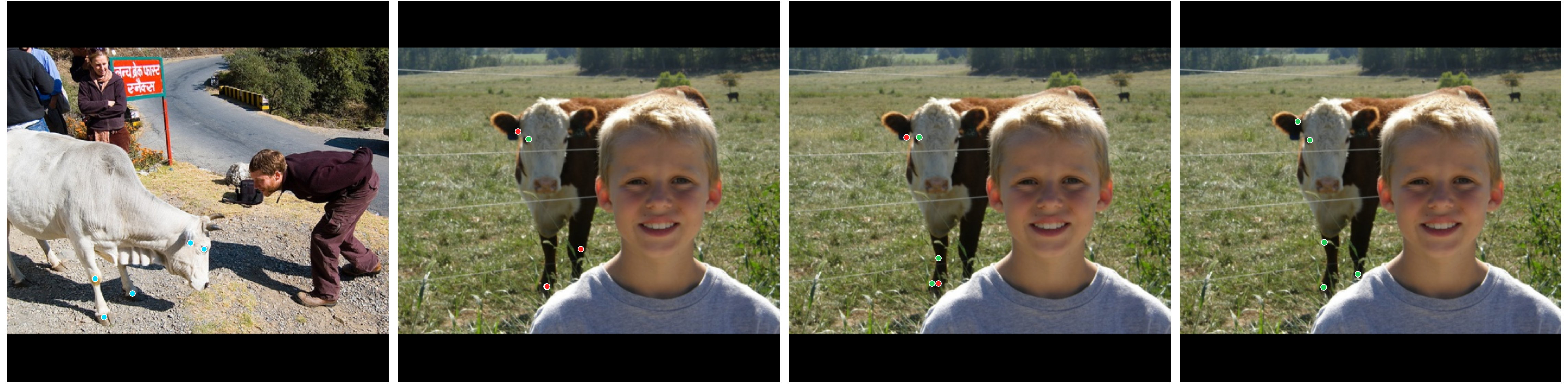}
\includegraphics[width=0.65\linewidth]{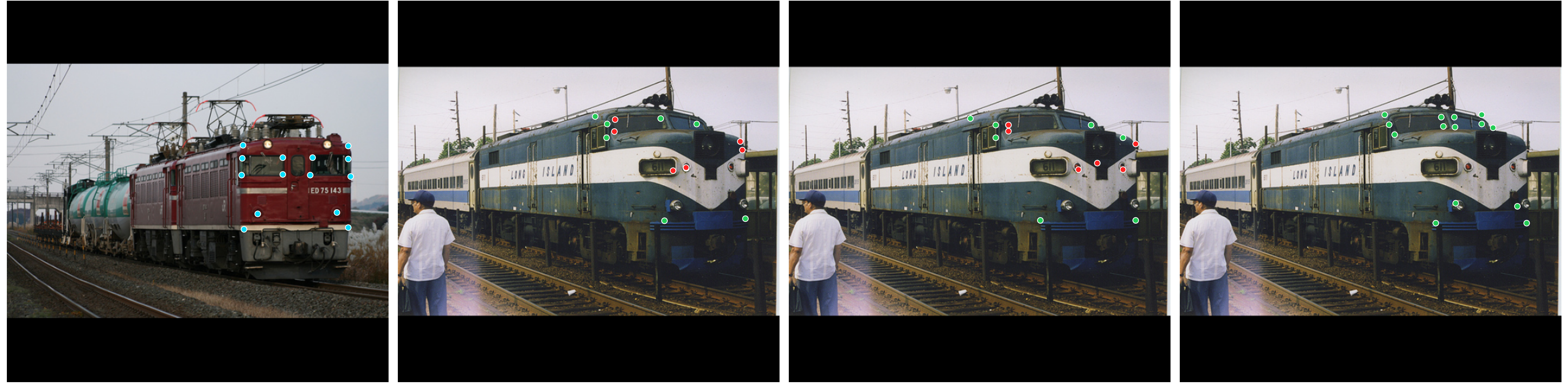}
\includegraphics[width=0.65\linewidth]{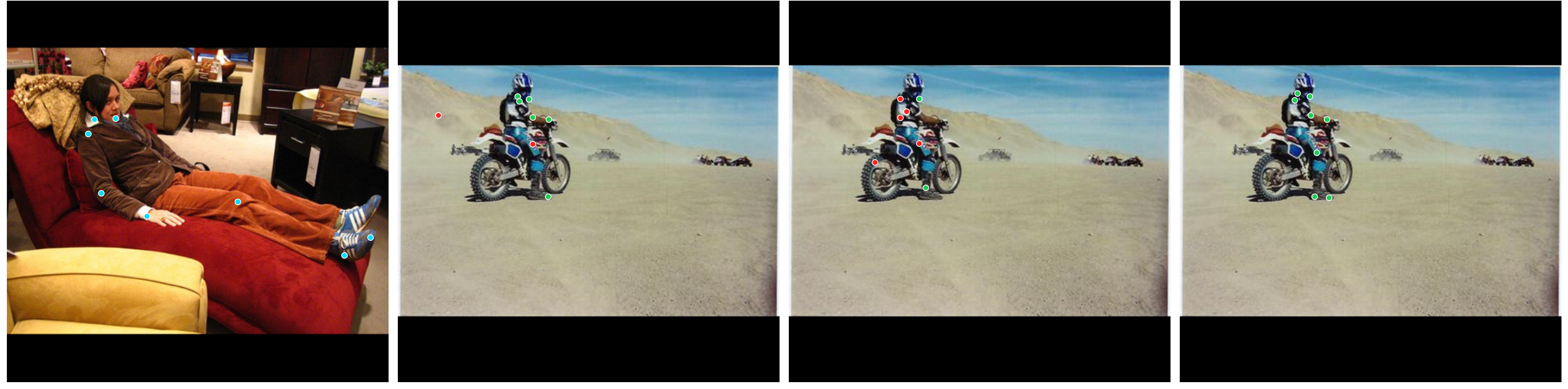}
\caption{Qualitative correspondence results on SPair-71k.}
\label{fig:qualitative}
\end{figure}

Figure~\ref{fig:qualitative} presents qualitative correspondence results on SPair-71k~\cite{min2019spair}, comparing DiTF~\cite{gan2025unleashing}, SD+DINO~\cite{zhang2023tale}, and DTM.
Green and red dots indicate correct and incorrect correspondences, respectively, under the PCK criterion with $\alpha_\text{bbox}=0.1$.
The methods differ in positional precision and in the number of correct matches.
These differences are especially noticeable for small and distinctive parts, such as beaks and noses.

\section{Limitations and Future Work}
\label{sec:limit}

\begin{figure}[htbp]
\centering
\includegraphics[width=0.85\linewidth]{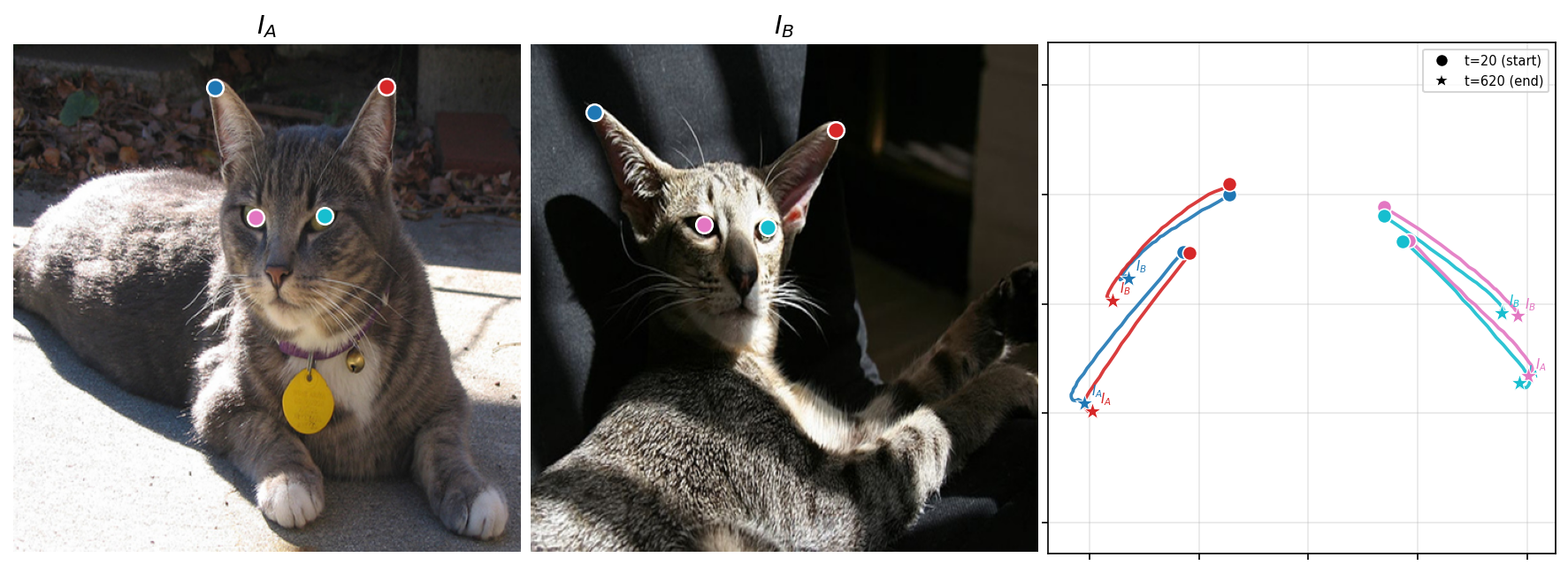}
\caption{Trajectory visualization for a cat image pair. Patches corresponding to symmetric parts (e.g., left/right eyes, left/right ears) exhibit similar trajectory patterns, illustrating the symmetric ambiguity challenge discussed in Section~\ref{sec:limit}.}
\label{fig:limit}
\end{figure}

This work has several limitations. First, DTM exhibits limitations for objects with bilateral symmetry. For such objects, semantically corresponding parts and their mirror counterparts share highly similar visual features, leading to similar trajectory patterns that make it difficult to distinguish correct correspondences from their symmetric counterparts (Figure~\ref{fig:limit}). Incorporating additional representations or learning strategies to resolve such symmetric ambiguity remains an open direction for future work. Second, while we empirically observe through PCA visualization that diffusion trajectories suggest semantic information, a theoretical explanation of how such temporal patterns give rise to meaningful representations remains lacking. Establishing a theoretical foundation for trajectory-based representations is an important direction for future research. Third, the process of extracting features across multiple timesteps and encoding them with Mamba incurs additional computational overhead. Exploring efficient trajectory encoding techniques or model compression methods remains an open problem for future work.

\section{Conclusion}

In this paper, we presented a novel perspective that treats diffusion representations as temporally structured trajectories, and proposed Diffusion Trajectory Modeling (DTM), a framework that leverages this view for semantic correspondence. Through our analysis, we showed that semantically corresponding spatial patches form similar trajectory patterns throughout the diffusion process, and demonstrated that the proposed method consistently outperforms single-timestep and simple aggregation-based representations. This work shows that the temporal structure inherent to the diffusion process can be directly exploited, presenting a new perspective for interpreting diffusion representations from a trajectory-centric viewpoint.

\bibliography{egbib}

\end{document}